\newcommand{\CLASSINPUTinnersidemargin}{0.67in} 
\newcommand{\CLASSINPUToutersidemargin}{0.67in} 
\newcommand{\CLASSINPUTtoptextmargin}{0.80in}   
\newcommand{\CLASSINPUTbottomtextmargin}{1.05in}
\documentclass[conference]{IEEEtran}
\IEEEoverridecommandlockouts

\usepackage{cite}
\usepackage{amsmath,amssymb,amsfonts}
\usepackage{graphicx}
\usepackage{textcomp}
\usepackage{xcolor}
\usepackage{algorithm}
\usepackage{tabularx}
\usepackage{algpseudocode}   
\usepackage{setspace}        
\usepackage{booktabs}  
\usepackage{array}
\usepackage{siunitx}
\usepackage{balance}
\def\BibTeX{{\rm B\kern-.05em{\sc i\kern-.025em b}\kern-.08em
    T\kern-.1667em\lower.7ex\hbox{E}\kern-.125emX}}
\begin{document}

\title{A Multi-Agent System for Building-Age Cohort Mapping to
Support Urban Energy Planning\\
\thanks{*corresponding author Kundan Thota}
}

\author{
\IEEEauthorblockN{Kundan Thota}
\IEEEauthorblockA{
\textit{Institute for Automation} \\
\textit{and Applied Informatics (IAI)} \\
\textit{Karlsruhe Institute of Technology (KIT)}\\
Karlsruhe, Germany \\
kundan.thota@kit.edu}
\and
\IEEEauthorblockN{Thorsten Schlachter}
\IEEEauthorblockA{
\textit{Institute for Automation} \\
\textit{and Applied Informatics (IAI)} \\
\textit{Karlsruhe Institute of Technology (KIT)}\\
Karlsruhe, Germany \\
thorsten.schlachter@kit.edu}
\and
\IEEEauthorblockN{Veit Hagenmeyer}
\IEEEauthorblockA{
\textit{Institute for Automation} \\
\textit{and Applied Informatics (IAI)} \\
\textit{Karlsruhe Institute of Technology (KIT)}\\
Karlsruhe, Germany \\
veit.hagenmeyer@kit.edu}
}

\maketitle

\begin{abstract}
Determining the age distribution of the urban building stock is crucial for sustainable municipal heat planning and upgrade prioritization. However, existing approaches often rely on datasets gathered via sensors or remote sensing techniques, leaving inconsistencies and gaps in data. We present a multi-agent LLM system comprising three key agents, the Zensus agent, the OSM agent, and the Monument agent, that fuse data from heterogeneous sources. A data orchestrator and harmonizer geocodes and deduplicates building imprints. Using this fused ground truth, we introduce BuildingAgeCNN, a satellite-only classifier based on a ConvNeXt backbone augmented with a Feature Pyramid Network (FPN), CoordConv spatial channels, and Squeeze-and-Excitation (SE) blocks. Under spatial cross-validation, BuildingAgeCNN attains an overall accuracy of 90.69\% but a modest macro-F1 of 67.25\%, reflecting strong class imbalance and persistent confusions between adjacent historical cohorts. To mitigate risk for planning applications, the address-to-prediction pipeline includes calibrated confidence estimates and flags low-confidence cases for manual review. This multi-agent LLM system not only assists in gathering structured data but also helps energy demand planners optimize district-heating networks and target low-carbon sustainable energy systems.
\end{abstract}

\section{Introduction}
The ongoing transition from high-carbon to sustainable green energy systems increasingly focuses on municipal heating networks as a primary means to decarbonize residential and commercial heating~\cite{Duchon24-platform-ecosystem}. The existing building stock in many countries of Europe has legacy construction practices, outdated rooftops, and inefficient heating systems~\cite{csoknyai2016}, thus resulting in greenhouse gas emissions. The age distribution of building stock is fundamentally an important characteristic for an energy planner to forecast future energy demand. The age of a building directly correlates with different parameters such as insulation levels, rooftop materials, type of heating systems, etc. Typically, the older buildings (e.g., pre-1919) consume more energy as their insulation levels are not optimized up to the mark, thus resulting in higher-per-square-meter heat losses \cite{tariku2023,aksoezen2015}. As a result, the energy-demand models require the age distribution of building stock to produce insights for low-carbon heating systems.

Traditional methods of gathering building age information happen in either of the two approaches: (1) official surveys or open-data portals, often providing incomplete coverage of construction-year metadata. (2) remote-sensing methods, where characteristics are extracted from street-level or satellite imagery, and the approximate age categories are inferred~\cite{rosser2019}. While both methods produce reasonable signals, neither alone offers comprehensive coverage. According to~\cite{maennig2011monument, destatis2018}, the monument registry has publicly available metadata for several federal states covering one million monuments, which is 5 - 7\% of the buildings in Germany. Often, this data is unstructured and available in different formats like PDFs, web pages, raw XML formats, etc. Gathering structured metadata from such heterogeneous sources would consume much manual work. Fortunately, LLMs can automatically parse vast, unstructured texts and easily generate structured output.

\begin{figure}[!t]
  \centering
  \includegraphics[width=\linewidth]{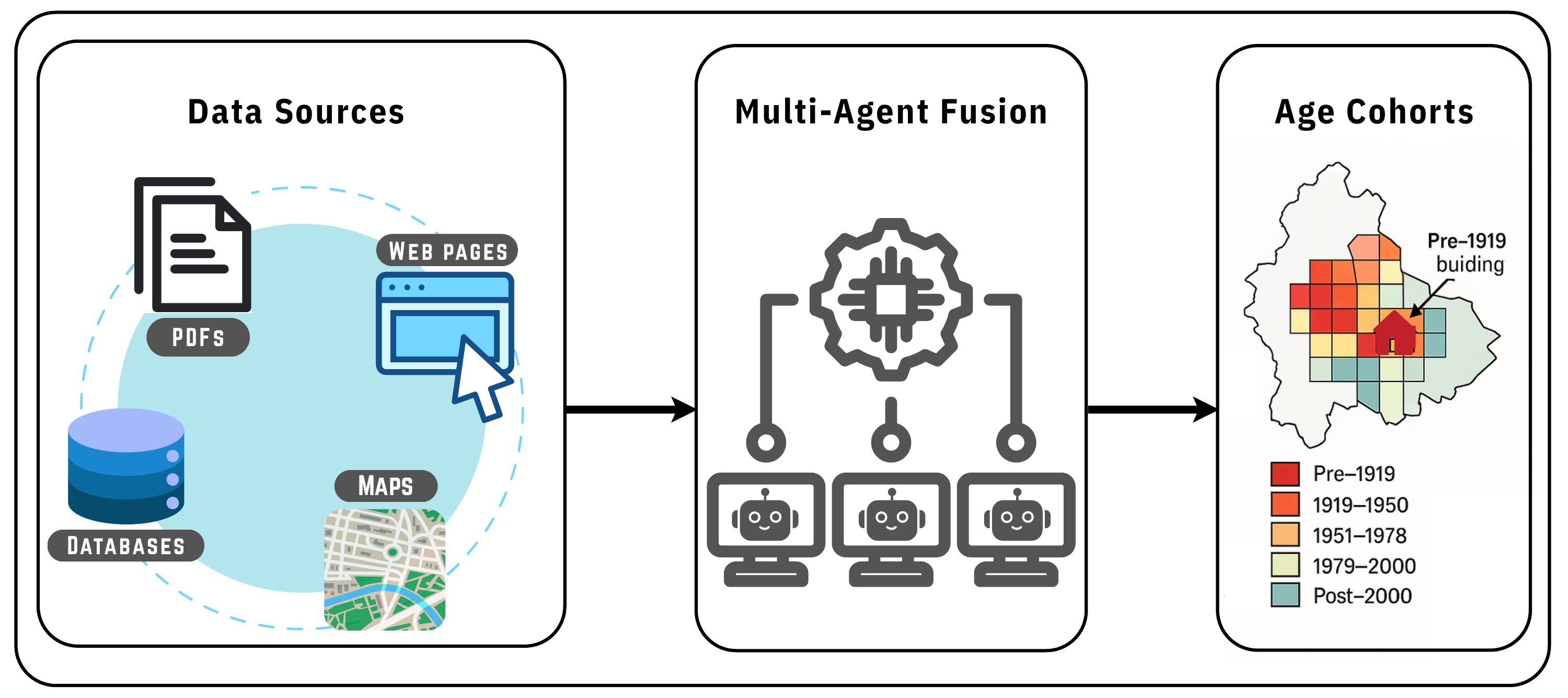}
  \caption{Outline: (1) Multiple data sources (databases, web pages, maps, and PDFs) are ingested into (2) LLM-driven agents to produce (3) geocoded age cohorts.}
  \label{fig:overview}
\end{figure}

In this paper, we propose a multi-agent LLM system that ingests data from heterogeneous sources. This architecture aims to extract the construction year from various sources such as the monument registry web source, OpenStreetMap (OSM) \cite{haklay2008openstreetmap}, and German census (Zensus 2011) \cite{lamla2010zensus}. This extracted data is then synchronized into a geocoded dataset for building addresses labeled by age cohorts, as shown in Fig.~\ref{fig:overview}. Within this architecture, individual agents are assigned distinct responsibilities to facilitate data formatting and synthesis. This ground truth provides a comprehensive training set for subsequent image-based classification.

In the second phase, we capture a high-resolution RGB satellite image of each building and train a CNN to classify the buildings into age cohorts. This model, hereafter referred to as BuildingAgeCNN, serves as the core part for building age estimation. This network learns from the captured high-resolution visual signatures, such as roof geometry, material reflectivity, and construction patterns. By deploying the classifier within the multi-agent architecture inference pipeline, the agent can find discrepancies in the classifier's confidence levels and flag the building for a manual review. This step helps the urban planner further evaluate the building from additional sources, yielding a fine-grained age cohort of building stock that supports accurate municipal energy planning.

Our main contributions are as follows:
\begin{itemize}
\item We design a multi-agent LLM system that gathers, standardizes, and orchestrates the building data from heterogeneous sources to construct a high-coverage, geocoded dataset.
\item We propose BuildingAgeCNN, a satellite-only classifier that combines a ConvNeXt backbone with a FPN, CoordConv spatial channels, and SE attention to improve multi-scale, spatially-aware features for building-age cohort classification.
\item We introduce an AgeCohort Agent within the inference pipeline to identify discrepancies in the classifier's confidence levels, flagging the building for a manual review.
 \end{itemize}

The paper is organized as follows: Section~\ref{sec:relatedwork} reviews related work on MAS for data fusion, CNN-based building‐age estimation, and building age-focused energy demand models. Section~\ref{sec:method} details the proposed architecture. Section~\ref{sec:results} shows the experimental results of three modules and an ablation study, while Section~\ref{sec:discussion} discusses the limitations and practical implementation and Section~\ref{sec:conclusion} concludes the paper and outlines the direction of future research.

\section{Related Work}
\label{sec:relatedwork}
Research on the building-age cohorts and their integration into municipal energy planning has grown significantly in recent years. In this section, we review the literature in three main areas: (1) applications of age-classified building inventories in urban energy planning, (2) building-age estimation methods, and (3) LLM-based multi-agent systems.

\subsection{Metadata for Urban Energy Models}
Often, energy models are calculated based on building-age data extracted from structured registers or open-data portals. Garbasevschi et al. estimate residential buildings age in North-Rhine Westphalia, Germany, using Random Forests on open spatial data and urban-morphology metrics, demonstrating the impact of spatial autocorrelation on heat-demand models~\cite{garbasevschi2021}. Ergan et al. incorporate construction year alongside features such as floor count and gross area into an XGBoost model; they capture the influence of older buildings, which generally exhibit higher energy demand, on energy‐demand behaviour \cite{yu2022}. Creutzig et al. and Reinhart \& Cerezo Davila implement workflows for bottom-up energy simulations, where building's age informs envelope U-values and ventilation rates~\cite{reinhart2016, nachtigall2023}. However, these methods either suffer from incomplete coverage or require manual schema alignment.

\subsection{Image-Based Building-Age Inference}
Benz et al. collect a 5307 street-view image dataset in Weimar, Germany, and train a convolutional ResNet classifier to predict the age cohort of the buildings with near-human accuracy \cite{benz2023}. Li et al. train a CNN on 1m RGB satellite patches labelled with OSM \texttt{building:year\_built} tags, achieving 85\% accuracy but facing sparse tag coverage \cite{kang2018}. Garbasevschi et al. also examine SIFT and HOG features for style-based inference, noting limited transferability across cities \cite{garbasevschi2021}. Our approach incorporates MAS-derived ground truth to train BuildingAgeCNN on high-resolution satellite patches, boosting the coverage and accuracy.

\subsection{LLM-Driven Multi-Agent systems}
Recent advances demonstrate the capability of LLM agents in data ingestion and tool orchestration. Yang et al. proposed LaMAS, a protocol for task decomposition and value generation in LLM-based multi-agent systems \cite{yang2024}. Huang et al. introduced AUTOSCRAPER, a two-stage LLM framework that progressively generates and synthesizes web scrapers across HTML structures \cite{huang2024}. Zhou et al. develop GMAC, using GPT-based semantic extraction to compress inter-agent communications by 53\% in MARL settings \cite{zhou2023}. To our knowledge, no prior work has applied LLM-driven agents to generate building-age labels from different formatted sources.

In summary, prior literature has advanced methods for age estimation, ranging from survey-based lookups to remote sensing. However, a scalable, agent-based data-fusion pipeline that consolidates data from multiple sources in monument registry documents remains an open challenge.

\section{Proposed Method}
\label{sec:method}
This section presents the pipeline for building-age cohort classification. It contains four key components, as shown in Fig.\ref{fig:architecture}: (1) a multi-agent system for data collection, (2) a Data orchestration, fusion and harmonization module for integrating and standardizing heterogeneous datasets, (3) a supervised learning BuildingAgeCNN, and (4) an inference module for real-time predictions.

\subsection{Multi‐Agent System for Data Collection}
\label{sec:fusion}
To construct a high-quality dataset for building-age cohorts, we developed a multi-agent system that collects data from heterogeneous public sources. The agents include:
\begin{figure*}[ht]
  \centering
  \includegraphics[width=0.85\textwidth]{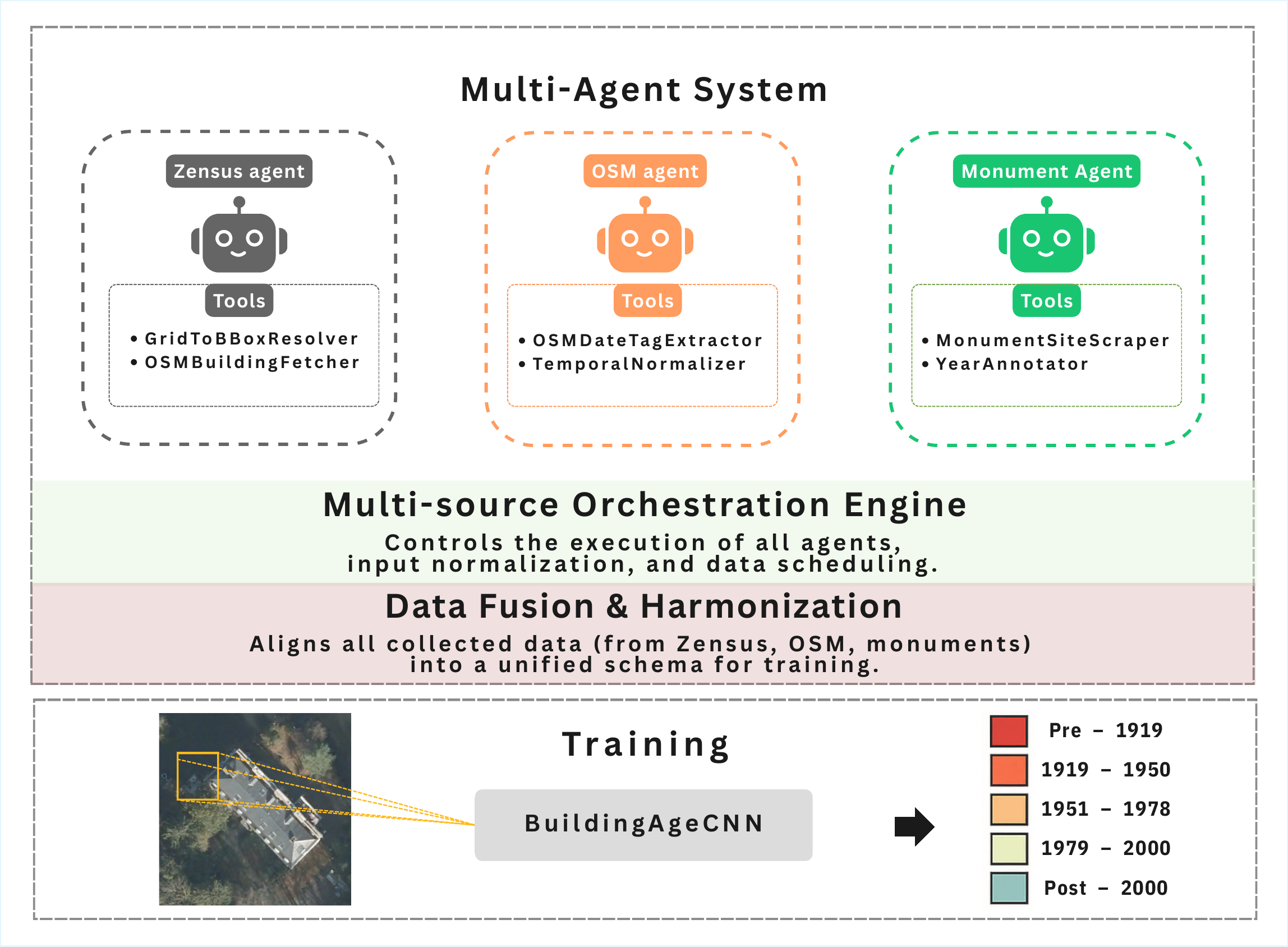}
  \caption{Overview of the proposed pipeline for building-age cohort mapping: A multi-agent system for building-age cohort mapping consists of three data‐collection agents (Zensus, OSM, Monument), then fed into a Data Orchestration, Fusion, and Harmonization that integrates and standardizes records into five age cohorts. The BuildingAgeCNN model (ConvNeXt + FPN + CoordConv + SE) is then trained on satellite imagery for supervised learning and inference.}
  \label{fig:architecture}
\end{figure*}

\subsubsection{Zensus agent}

The Zensus agent is responsible for enriching the building dataset with age cohort labels derived from the official German census. Each building is related to a 100-meter grid area within the survey, identified by a unique ID. In order to fulfil the tasks, the Zensus agent utilizes the following tools:
\begin{itemize}
\item \textbf{GridToBBoxResolver Tool}: Parses the grid ID and converts the coordinates from EPSG:3035 to EPSG:4326. This produces the bounding box for a 100m x 100m tile.
\item \textbf{OSMBuildingFetcher Tool}: Queries OSM via overpass API~\cite{olbricht2011overpass}  to retrieve all building footprints within the bounding box.
\end{itemize}

\subsubsection{OSM agent}

The OSM agent extracts building-level construction year data from OpenStreetMap. The agent works over the Overpass API and invokes a set of modular tools as follows:

\begin{itemize}
        \item \textbf{OSMDateTagExtractor Tool}: Extract all building polygons that include either the \texttt{building:start\_date} or \texttt{building:year\_built} tag, as these tags provide direct or approximate building's construction year.

     \item \textbf{TemporalNormalizer Tool}: An LLM is used to convert year descriptions into estimated construction periods in cases where the year information is historical. For example, "early 19C" is mapped to pre-1919, and "mid-20C" is equated to 1951–1978. This process makes the dataset's temporal details more precise without losing its interpretability.
\end{itemize}
\subsubsection{Monument agent}

The Monument agent extracts building information from the historical monuments listed on the Monument web portal~\cite{Aachen2025_HistoricPreservation}, which includes semi-structured data such as building descriptions and addresses that are not consistently formatted. The agent uses the following tools:

\begin{itemize}
    \item \textbf{MonumentSiteScraper Tool}: Fetches the HTML content of each monument entry using HTTP requests. The page is parsed with BeautifulSoup to extract content from specific DOM regions, such as addresses and building descriptions.

    \item \textbf{YearAnnotator Tool}: Uses a well-structured prompt to interpret construction descriptions and map them to age cohorts using a structured prompt. The LLM is instructed to return a JSON object with a single field: \texttt{construction\_year}.
\end{itemize}

\subsection{Data orchestration, fusion and harmonization}
\label{sec:data_orch}

The Monument, Zensus, and OSM agents do not exchange messages with each other and do not operate in a turn-based or dialog-style interaction. Instead, all agents execute independently and in parallel on their respective data sources. 
All agents operate under a shared LLM input--output contract, where raw text or metadata and a fixed, closed cohort set are provided as input, and a validated \texttt{JSON}-only structure \texttt{year\_raw} is returned.

A lightweight Orchestrator coordinate these agents and maps building records to a standardized geocoded schema.
\[
 {\textit{lat},\ \textit{lon},\ \textit{year\_raw},\  \textit{source}}
\]
The Data Fusion module forms a combined dataset by concatenating the geocoded outputs from three agents. The Data Harmonizer identifies the duplicate records using identical \texttt{(lat,lon)}. Then it selects a construction year for each cluster of duplicates by a deterministic priority rule that prefers the most definitive source:
\[
\text{chosen\_year} \;=\;
\begin{cases}
\text{from monument registry}, & \text{if present;}\\[4pt]
\text{from census},  & \text{else if present;}\\[4pt]
\text{from openstreetmap },     & \text{else if present;}\\[4pt]
\text{null}          & \text{otherwise.}
\end{cases}
\]
If no source provides a construction year, the clustered record is dropped. To better understand this entire pipeline, the compact procedure is summarized in Algorithm~\ref{alg:orchestrator}. 

\begin{algorithm}[!ht]
\caption{Orchestration, Fusion and Harmonization}
\label{alg:orchestrator}
\centering
\begin{minipage}{0.8\columnwidth}
\raggedright\small
\begin{spacing}{1.25}   
\begin{algorithmic}
\Procedure{Unified Dataset}{}
  \Statex \textbf{Orchestration:}
  \State \quad Run concurrently $\rightarrow$ \{\textsc{Monument}, \textsc{OSM}, \textsc{Zensus}\}
  \State \quad $M \gets$ \textsc{Monument\_Agent}()
  \State \quad $O \gets$ \textsc{OSM\_Agent}()
  \State \quad $Z \gets$ \textsc{Zensus\_Agent}()
  \Statex

  \Statex \textbf{Data fusion:}
  \State \quad $R \gets M \cup O \cup Z$ \Comment{concatenate results}
  \Statex

  \Statex \textbf{Data harmonization:}
  \State \quad $G \gets \text{group\_by}(R,\ \text{key}=(lat,lon))$
  \ForAll{group $g \in G$}
    \State \quad $chosen\_year \gets \text{first\_available}(g, [\text{Monument},\text{Zensus},\text{OSM}])$
    \If{$chosen\_year \ne \text{null}$}
      \State \quad append (lat, lon, chosen\_year, chosen\_source) to \texttt{fused\_table}
    \EndIf
  \EndFor

  \State \Return \texttt{fused\_table}
\EndProcedure
\end{algorithmic}
\end{spacing}
\end{minipage}
\end{algorithm}

The normalized construction year is then mapped into one of five age cohorts pre-1919, 1919–1950, 1951–1978, 1979–2000, post-2000).

\subsection{BuildingAgeCNN for Age Cohort Mapping}
\label{sec:cnn}

The proposed training pipeline utilizes BuildingAgeCNN, a hybrid CNN architecture developed to categorize buildings into age cohorts using satellite imagery. We adopt five recognized age cohorts (pre--1919, 1919--1950, 1951--1978, 1979--2000, post--2000) based on significant material and regulatory shifts \cite{IOER2025MaterialComposition, loga2016tabulaure}. A pre-trained ConvNeXt~\cite{liu2022convnet} model is utilized as the feature extractor. To incorporate semantic information at different scales, we used FPN~\cite{lin2017feature} that takes input from 'layer3' and 'layer4'. 

Then, this FPN feature map is passed through a CoordConv layer where the center coordinates \(X\) and \(Y\in\mathbb{R}^{1\times H\times W}\) are concatenated with the feature vector, allowing each spatial location to carry its own normalized position before filtering:
\begin{equation}
f_{\mathrm{coord}}
= \mathrm{Conv2D}\bigl(\,[f_{\mathrm{fpn}};\,X;\,Y]\,\bigr)
\;\in\;\mathbb{R}^{C'\times H\times W}.
\end{equation}
 By providing \((X,Y)\) information at every pixel, the network easily distinguishes the crucial location-dependent patterns, leading to faster convergence and improved performance.

Subsequently, a SE \cite{hu2018squeeze} block is applied to refine feature channels based on global context:
\begin{equation}
f_{\text{se}} = f_{\text{coord}} \cdot \sigma(W_2 \cdot \text{ReLU}(W_1 \cdot \text{GAP}(f_{\text{coord}})))
\end{equation}
where \( \text{GAP}(\cdot) \) denotes global average pooling, \( W_1 \) and \( W_2 \) are learnable weights, and \( \sigma \) is the sigmoid activation function.

Lastly, the feature map is passed through a convolutional classifier head that includes a \(3 \times 3\) convolution, ReLU activation, adaptive average pooling, and fully connected layers to classify a building into one of the age cohorts.

\subsection{Inference module}
\label{sec:inference}

\begin{figure}[!b]
  \centering
  \includegraphics[width=\columnwidth]{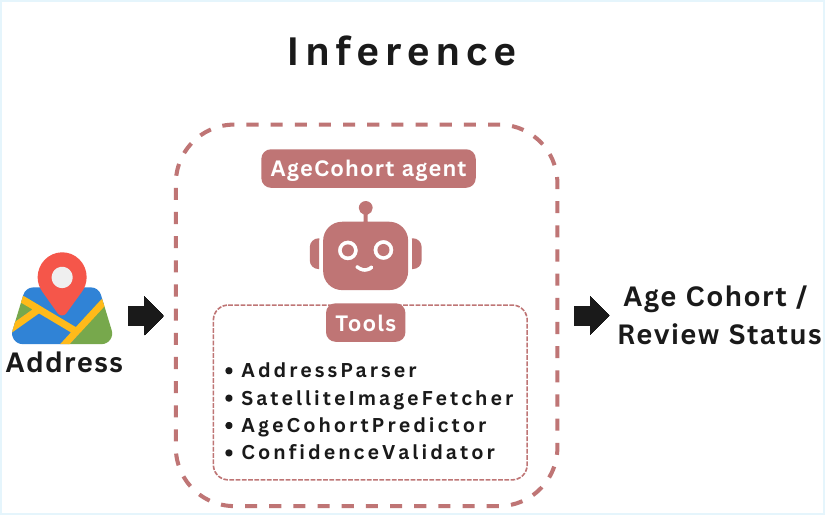}
  \caption{Inference pipeline of the AgeCohort agent: address parsing, geocoding, satellite-tile retrieval, BuildingAgeCNN inference, and confidence-based flagging for manual review.}
  \label{fig:inference}
\end{figure}

During inference, the AgeCohort agent uses four modular tools, as shown in Fig.~\ref{fig:inference}, to convert the input address into a building‐age cohort or flag a building with a low confidence level for a manual review.

\begin{itemize}
  \item \textbf{AddressParser Tool:} Extracts street name, house number, and city from the input text using an LLM system with a well-defined system prompt.
  \item \textbf{SatelliteImageFetcher Tool:} Geocodes the address to geo-coordinates and fetches the fixed‐resolution RGB satellite tile centered on the building footprint.  
  \item \textbf{AgeCohortPredictor Tool:} Loads the trained BuildingAgeCNN model and computes softmax probabilities over the five age cohorts.  
  \item \textbf{ConfidenceValidator Tool:} Compares the top‐1 probability \(p_{\max}\) against a preset threshold \(\tau\). If \(p_{\max} \geq \tau\), the predicted cohort is returned; otherwise, the building is flagged for manual review:
  \[
    \text{Decision} =
    \begin{cases}
      \arg\max_i p_i, & \text{if } p_{\max} \ge \tau,\\
      \texttt{FLAGGED}, & \text{if } p_{\max} < \tau.
    \end{cases}
  \]
\end{itemize}

This workflow allows energy planners to retrieve age estimates using only an address input, streamlining large-scale building-age cohorts to support urban energy planning.

\section{Experimental Results}
\label{sec:results}

\subsection{Dataset coverage analysis}
\label{sec:coverage}

To evaluate the effectiveness of our multi-agent data-fusion pipeline, we compare the fused dataset with the full set of OSM building footprints for Aachen, Germany, which provides a publicly available registry of monuments. Table~\ref{tab:coverage_sources} reports overall coverage: the fused dataset contains 15,336 buildings with age-cohort labels, covering 21.17\% of the 72,446 OSM footprints in Aachen.

This reduced coverage primarily reflects limitations in data availability rather than pipeline failure. For the majority of OSM building footprints in Aachen, construction-period information is not publicly available in structured or machine-readable form and is therefore excluded from the fused dataset. In addition, buildings with missing, ambiguous, or conflicting age signals are conservatively filtered to ensure label reliability. Consequently, the fused dataset represents a high-confidence labeled subset of the building stock, rather than a complete census of all buildings.

\begin{table}[!b]
\caption{Coverage of age-labeled buildings in Aachen (sources and fused output).}
\begin{center}
\begin{tabular}{|l|c|c|}
\hline
\textbf{Data Source} & \multicolumn{2}{|c|}{\textbf{Coverage Statistics}} \\
\cline{2-3}
 & \textbf{Building Count} & \textbf{Coverage (\%)} \\
\hline
OSM total buildings & 72,446 & 100.00 \\
Fused dataset       & 15,336 & 21.17  \\
\hline
\end{tabular}
\label{tab:coverage_sources}
\end{center}
\end{table}

\begin{table}[!b]
\caption{Age-cohort distribution in the fused dataset.}
\begin{center}
\begin{tabular}{|l|c|c|}
\hline
\textbf{Cohort} & \multicolumn{2}{|c|}{\textbf{Dataset Statistics}} \\
\cline{2-3}
 & \textbf{Count} & \textbf{Share (\%)} \\
\hline
pre--1919      & 2,722  & 17.76 \\
1919--1950     & 1,114  & 7.26  \\
1951--1978     & 10,212 & 66.60 \\
1979--2000     & 892   & 5.82  \\
post--2000     & 396   & 2.58  \\
\hline
\textbf{Total (fused)} & \textbf{15,336} & \textbf{100.00} \\
\hline
\end{tabular}
\label{tab:cohort_dist}
\end{center}
\end{table}

Table~\ref{tab:cohort_dist} breaks down the fused set by age cohort. The largest share (66.60\%) falls in the 1951–1978 cohort, followed by pre-1919 (17.76\%), 1919–1950 (7.26\%), 1979–2000 (5.82\%), and post-2000 (2.58\%). This stretch toward mid-20th-century building stock has direct implications for municipal heat planners to prioritize the 1951–1978 cohort, given its large contribution to the stock. 

\subsection{Roof‐Type Progression by Age Cohort}

\begin{figure}[!t]
  \centering
  \includegraphics[width=\columnwidth]{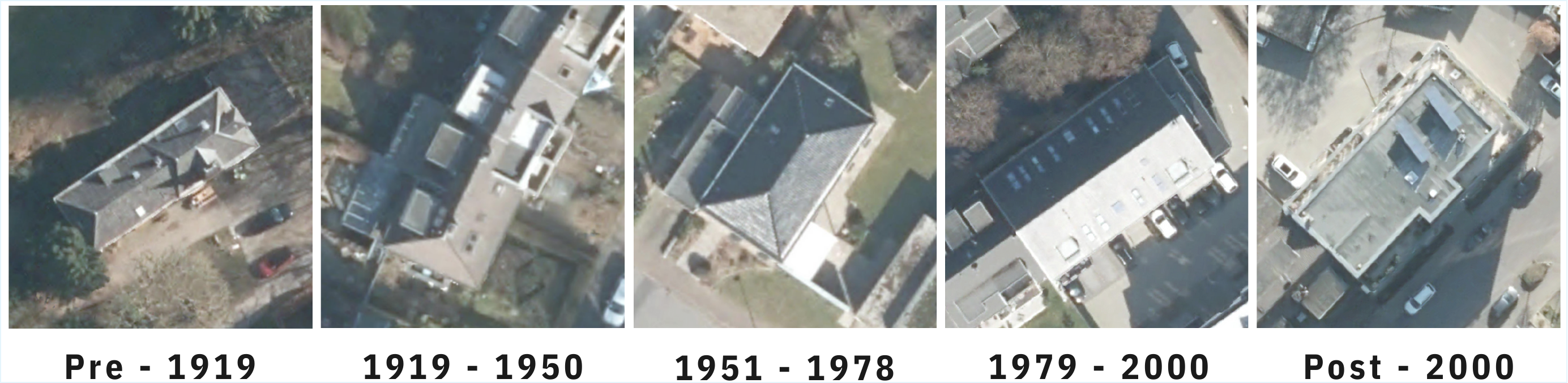}
  \caption{Representative roof types across age cohorts (left to right: pre-1919, 1919–1950, 1951–1978, 1979–2000, post-2000).}
  \label{fig:roof_types}
\end{figure}

Fig.~\ref{fig:roof_types} displays rooftop profiles for each of the five building‐age cohorts. A clear progression in the roof architecture is evident that the pre-1919 building is dominated by steep and tile‐covered gabled roofs, while the 1919–1950 cohort shows a lower‐slope roof with metal sheeting. From 1951–1978 onward, low-slope roofs acquired importance, and by the 1979–2000 cohort, many buildings shifted toward flat-roof designs. In the post-2000 cohort, flat roofs are equipped with solar PV panels. The shift from traditional pitched roofs to modern flat roofs incorporating energy-efficient features gives clear visual indicators that BuildingAgeCNN uses to differentiate building-age cohorts.

\subsection{BuildingAgeCNN experimental setup}
\label{sec:cnn_experiments}

This section outlines the experimental setup for evaluating the proposed BuildingAgeCNN with different CNN/ViT baselines. The experiments focus on robust spatial validation, ablations, and standardized training for fair comparison.

\subsubsection{Dataset and Spatial preprocessing}
\label{sec:dataset_preproc}
The dataset contains building-age cohorts, each linked to an RGB satellite tile of size $224\times224$. All images are resized and normalized using ImageNet mean and standard deviation. Data augmentation includes random horizontal flips, rotations within \( \pm 10^{\circ} \).

\begin{figure}[!t]
  \centering
  \includegraphics[width=0.8\columnwidth]{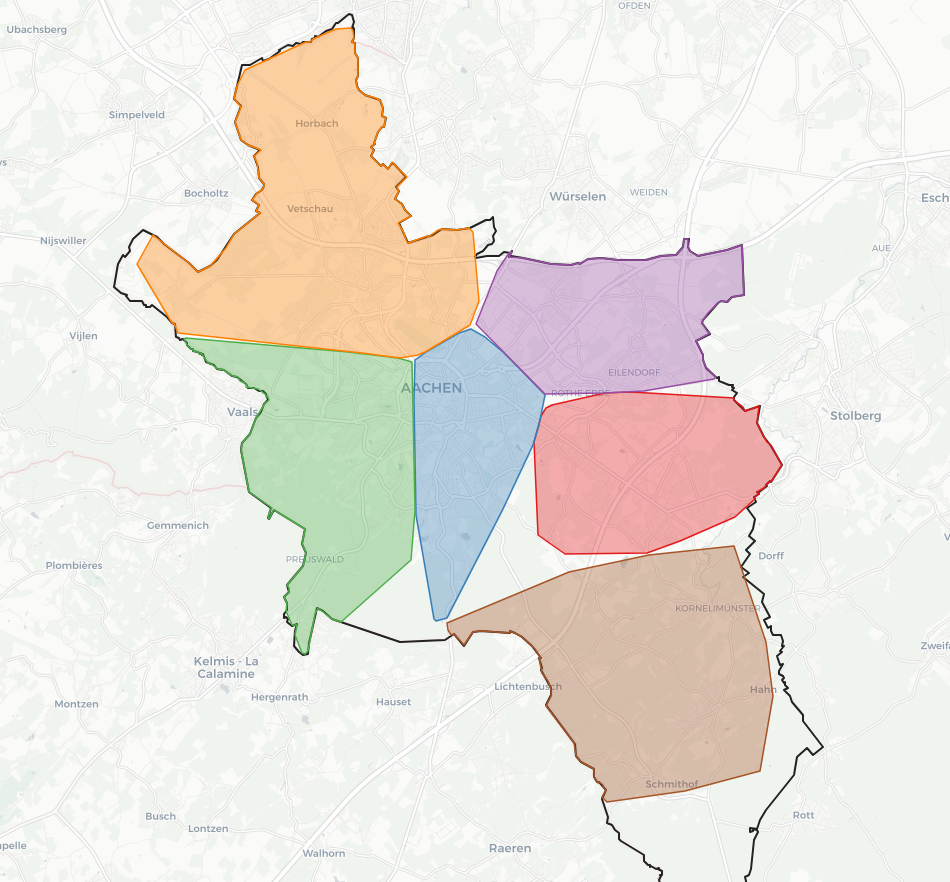}
  \caption{Data combined into six spatially contiguous folds, with each fold visualized as a region polygon.}
  \label{fig:aachen}
\end{figure}

To ensure spatial independence, we employed: \\
\textbf{Spatial-Cluster cross-validation}: KMeans clustering with K=6 on standardized coordinates (\texttt{lat},\texttt{lon}) and performed the 6-fold cross-validation, ensuring that no spatially adjacent buildings appear in different folds as shown in the Fig.~\ref{fig:aachen}.

\subsubsection{Model Architecture and Training}
The BuildingAgeCNN architecture builds upon the convolutional backbones, progressively enhanced with multi-scale and spatially aware channels. 
\begin{itemize}
  \item \textbf{Feature Pyramid Networks (FPN):} for multi-scale fusion,
  \item \textbf{CoordConv:} for spatial awareness, and
  \item \textbf{Squeeze-and-Excitation (SE):} for channel recalibration.
\end{itemize}

All models are initialized with ImageNet weights. By default, backbone parameters remain frozen while classifier heads are trained, with staged unfreezing explored in ablations.
Training uses cross-entropy loss with label smoothing (\(0.05\)), AdamW optimizer (\(\text{LR}=3\times10^{-4}\), weight decay \(=1\times10^{-4}\)), and cosine annealing scheduling over 30 epochs with weighted sampling for class imbalance. 

\subsubsection{Ablation study}
We conduct an ablation study to quantify the contribution of different components. Starting with a baseline, we incrementally added FPN,  CoordConv, and SE blocks. We benchmark several backbones such as ResNet-50~\cite{koonce2021resnet}, MobileNetV3~\cite{howard2019searching}, EfficientNet-B0~\cite{tan2019efficientnet}, ConvNeXt, a lightweight custom CNN and the results are summarized in Table~\ref{tab:ablation}. Adding CoordConv and SE blocks significantly improved performance, while FPN contributed to better feature aggregation, indicating that these modules are architecture complementary.

\begin{table*}[!t]
\caption{Ablation results on the building-age benchmark.}
\setlength{\tabcolsep}{8pt} \renewcommand{\arraystretch}{1.0}
\centering
\begin{tabular}{|l|l|c|c|c|}
\hline
\textbf{Model} & \textbf{Variant} & \multicolumn{3}{|c|}{\textbf{Metrics}} \\
\cline{3-5}
 &  & \textbf{Accuracy (\%)} & \textbf{Macro-F1} (\%)& \textbf{Params (M)} \\
\hline
ConvNeXt~\cite{liu2022convnet} & Baseline & \(82.27 \pm 5.55\) & \(63.71 \pm 5.57\) & 27.82 \\
 & + FPN & \(89.06 \pm 5.43\) & \(64.36 \pm 5.35\) & 30.48 \\
 & + FPN + CoordConv & \(89.27 \pm 3.83\) & \(66.18 \pm 4.72\) & 31.08 \\
 & + FPN + CoordConv + SE & \(\mathbf{90.69} \pm \mathbf{4.22}\) & \(\mathbf{67.25} \pm \mathbf{5.52}\) & 31.09 \\
\hline
ResNet-50~\cite{koonce2021resnet} & Baseline & \(76.07 \pm 3.98\) & \(56.93 \pm 3.81\) & 25.60 \\
 & + FPN & \(78.94 \pm 3.66\) & \(57.23 \pm 6.32\) & 28.81 \\
 & + FPN + CoordConv & \(79.28 \pm 4.12\) & \(57.43 \pm 3.87\) & 30.12 \\
 & + FPN + CoordConv + SE & \(80.98 \pm 3.22\) & \(57.48 \pm 4.12\) & 30.13 \\
\hline
MobileNetV3~\cite{howard2019searching} & Baseline & \(80.84 \pm 6.62\) & \(59.40 \pm 5.25\) & 4.21 \\
 & + FPN & \(82.28 \pm 3.95\) & \(59.63 \pm 5.01\) & 5.78 \\
 & + FPN + CoordConv & \(83.40 \pm 7.10\) & \(60.41 \pm 6.32\) & 6.38 \\
 & + FPN + CoordConv + SE & \(84.51 \pm 6.00\) & \(62.38 \pm 4.97\) & 6.39 \\
\hline
EfficientNet-B0~\cite{tan2019efficientnet} & Baseline & \(72.21 \pm 0.16\) & \(51.77 \pm 0.05\) & 4.01 \\
 & + FPN & \(79.32 \pm 0.42\) & \(58.31 \pm 0.42\) & 7.07 \\
 & + FPN + CoordConv & \(81.23 \pm 0.22\) & \(59.01 \pm 0.91\) & 7.66 \\
 & + FPN + CoordConv + SE & \(83.28 \pm 0.38\) & \(62.69 \pm 0.41\) & 7.67 \\
\hline
Simple CNN & Baseline & \(68.90 \pm 0.15\) & \(49.18 \pm 0.11\) & 0.42 \\
 & + FPN & \(69.10 \pm 0.11\) & \(49.31 \pm 0.21\) & 2.66 \\
 & + FPN + CoordConv & \(70.92 \pm 0.28\) & \(50.10 \pm 0.32\) & 3.26 \\
 & + FPN + CoordConv + SE & \(73.10 \pm 0.43\) & \(51.21 \pm 0.21\) & 3.26 \\
\hline
\end{tabular}
\label{tab:ablation}
\end{table*}

\subsubsection{Classification performance}
\label{sec:classification_perform}
While overall accuracy is strong, Table~\ref{tab:classification} reveals
1919--1950 and post--2000 exhibit notably lower accuracy/macro-F1 than other classes. 
We attribute this to (i) \textbf{class imbalance} that biases the decision boundary towards the 1951--1978, (ii) \textbf{visual similarity} in roof materials, building footprints, and urban morphology between adjacent-era cohorts, and (iii) \textbf{renovation effects} that obscure temporal indicators in high-resolution satellite imagery.

\begin{table}[!t]
\caption{Per-class performance: accuracy and F1 by age cohort.}
\begin{center}
\begin{tabular}{|l|c|c|}
\hline
\textbf{Cohort} & \multicolumn{2}{|c|}{\textbf{Metrics}} \\
\cline{2-3}
 & \textbf{Accuracy (\%)} & \textbf{F1 (\%)} \\
\hline
pre--1919   & 81.96 & 88.84 \\
1919--1950  & 24.22 & 35.71 \\
1951--1978  & 96.38 & 89.95 \\
1979--2000  & 80.64 & 86.19 \\
post--2000  & 31.25 & 35.31 \\
\hline
\end{tabular}
\label{tab:classification}
\end{center}
\end{table}
\begin{figure}[!b]
  \centering
  \includegraphics[width=\columnwidth]{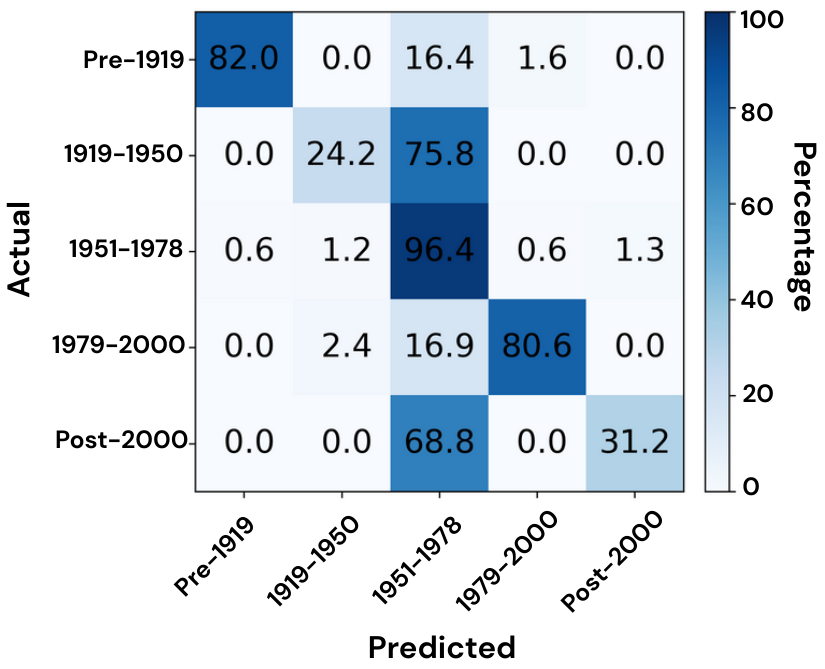}
  \caption{Row-normalized confusion matrix for five cohorts.}
  \label{fig:confusion}
\end{figure}

Fig.~\ref{fig:confusion} shows that 1951--1978 attains the highest true-positive rate (96.4\%), with limited leakage to neighboring cohorts. Pre--1919 is mostly correct (82.0\%) but exhibits spillover to 1951--1978 (16.4\%), suggesting texture/footprint similarities learned at mid-century scales. 
1919--1950 is frequently misclassified as 1951--1978 (75.8\%), and post--2000 is also often predicted as 1951--1978 (68.8\%). 
This systematic pull toward 1951--1978 aligns with the class-imbalance hypothesis and the visual proximity of adjacent-era construction.

\subsection{Inference Pipeline Demonstration}

\begin{table}[!t]
\caption{Flagged predictions at different confidence thresholds.}
\begin{center}
\begin{tabular}{|c|c|c|}
\hline
\textbf{Confidence Threshold} & \multicolumn{2}{|c|}{\textbf{Flagged Predictions}} \\
\cline{2-3}
 & \textbf{Count} & \textbf{Share (\%)} \\
\hline
0.65 & 128 & 18.28 \\
0.70 & 172 & 24.57 \\
0.75 & 259 & 37.00 \\
\hline
\end{tabular}
\label{tab:inference_flags}
\end{center}
\end{table}

We evaluated the AgeCohort agent on approximately 700 unseen building addresses in Aachen. Using three confidence‐threshold settings, the agent flags a varying number of predictions for manual review.

As shown in Table~\ref{tab:inference_flags}, raising the acceptance threshold increases the number of low-confidence cases requiring manual review from 18.28\% at \(\tau=0.65\) to 37.00\% at \(\tau=0.75\). While we do not have access to the records of the flagged buildings, a visual inspection of a few examples indicates that they have undergone renovations, like new roofing and solar panels. These often coincide with misclassified cases since they confuse the original architectural features our model relies on. Higher thresholds reduce false predictions but increase manual review, while lower thresholds maximize automation but increase the number of uncertain cases.

\subsection{Application to Energy Simulation}
To assess the practical relevance of the predicted age cohorts, we compare the cohort outputs against the representative U-values from the TABULA database (Table~\ref{tab:cohort_uvals})~\cite{ballarini2014use}. 
These values summarize typical thermal transmittances for roofs, upper ceilings, walls, and floors by construction period. 
The results highlight a consistent downward trend in U-values across cohorts, confirming that newer buildings generally exhibit improved thermal performance. 
This alignment suggests that the BuildingAgeCNN classifications could support downstream energy simulations or retrofit planning, where cohort-based U-values provide a first-order approximation of building envelope efficiency.

\begin{table}[!t]
\caption{Cohort-averaged U-values (W/m\textsuperscript{2}K) for envelope components from TABULA reference data.}
\begin{center}
\begin{tabular}{|l|c|c|c|c|}
\hline
\textbf{Cohort} & \multicolumn{4}{|c|}{\textbf{Thermal Transmittance (U-values)}} \\
\cline{2-5}
 & \textbf{Roof} & \textbf{Upper Ceiling} & \textbf{Wall} & \textbf{Floor} \\
\hline
pre--1919      & 1.95 & 1.00 & 2.10 & 2.05 \\
1919--1950     & 1.54 & 1.06 & 1.41 & 1.41 \\
1951--1978     & 0.75 & 0.96 & 1.05 & 1.29 \\
1979--2000     & 0.42 & 0.39 & 0.52 & 0.62 \\
post--2000     & 0.24 & 0.25 & 0.28 & 0.33 \\
\hline
\end{tabular}
\label{tab:cohort_uvals}
\end{center}
\end{table}

\section{Discussion}
\label{sec:discussion}

\textbf{Practical implication for planners: } Age cohorts enable immediate plug-in to building stock simulators for energy consumption. Even imperfect cohort labels materially improve district-heating sizing and prioritization compared to having no age information, where planners can (i) run sensitivity studies by cohort, (ii) target manual surveys where the model flags uncertainty, and (iii) prioritize measurement campaigns for cohorts with high heat-loss potential.
\textbf{Failure causes:} The confusion matrix shows two dominant error patterns: (1) adjacent-period confusions (architectural similarity between neighboring cohorts) and (2) majority-collapse into the 1951–1978 class (class imbalance). Both reduce macro-F1 despite high accuracy.

\section{Conclusion}
\label{sec:conclusion}
This paper introduces an LLM-driven multi-agent pipeline that combines German Census 2011, OpenStreetMap and Monument-registry sources into a unified, geocoded dataset and uses it to supervise a satellite-only classifier,  BuildingAgeCNN (ConvNeXt backbone + FPN + CoordConv + SE). Our best model achieves an average accuracy of 90.69\% on spatial-cluster cross-validation and a macro-F1 score of 67.25\%. Ablation experiments show that FPN, CoordConv and SE blocks each add complementary value: FPN improves multi-scale aggregation, CoordConv encodes spatial priors, and SE improves channel-wise feature selection. The inference pipeline converts user addresses to cohort predictions, applies a confidence threshold, and flags low-confidence cases for human review, making the system practical for planner workflows while exposing where automation should be augmented by experts.

Key limitations include prominent class imbalance, confusion between visually similar neighbouring cohorts, visual changes due to renovations (e.g., new roof or solar panels), and occlusions that degrade reliability. To address these issues in future, we will (1) incorporate active-learning to acquire targeted labels for underrepresented cohorts, (2) perform cross-city transfer and domain-adaptation experiments to evaluate generalization, (3) extend modalities (street-view, 3D imagery) to reduce visual ambiguity, and link cohort outputs to energy-performance archetypes for direct planner use. We will also publish provenance metadata, prompts, and code to improve reproducibility.

Overall, the pipeline demonstrates a scalable path from heterogeneous registries to actionable age-cohort maps that can support sustainable urban heat planning. With the planned improvements, we expect substantial gains in per-class performance and greater utility for energy planners.
\section*{Acknowledgments}
This work was carried out within the project NEED~\cite{Duchon24-platform-ecosystem} and was supported by BMWE under Grant No. 03EN3077J. Portions of the manuscript text (e.g., wording suggestions and Grammar) were drafted with ChatGPT's Assistance. The authors reviewed, edited, and take full responsibility for all content.

\balance
\bibliographystyle{unsrt}
\bibliography{bibliography}

\end{document}